# WPS-Dataset: A benchmark for wood plate segmentation in bark removal processing

Rijun Wang [1,2*], Guanghao Zhang[1,2], Fulong Liang [1,2], Bo WANG[2,3*], Xiangwei Mou [1,2], and Yesheng Chen[1,2], Peng Sun[4], Canjin Wang[5*]

1 School of Teachers College for Vocational and Technical Education, Guangxi Normal University, Guilin 541004, China
2 Key Laboratory of AI and Information Processing, Hechi University, Yizhou 546300, China
3 School of Artificial Intelligence and Smart Manufacturing, Hechi University, Yizhou 546300, China
4 School of Information and Communication Engineering, North University of China, Taiyuan 030051, China
5 State Key Laboratory of Media Convergence Production Technology and Systems & Xinhua Zhiyun Technology Co., Ltd., Hangzhou 310000, China
* Correspondence: 05041@hcnu.edu.cn; wangrijun1982@126.com; wangcanjin@shuwen.com

**Abstract**：Using deep learning methods is a promising approach to improving bark removal efficiency and enhancing the quality of wood products. However, the lack of publicly available datasets for wood plate segmentation in bark removal processing poses challenges for researchers in this field. To address this issue, a benchmark for wood plate segmentation in bark removal processing named WPS-dataset is proposed in this study, which consists of 4863 images. We designed an image acquisition device and assembled it on a bark removal equipment to capture images in real industrial settings. We evaluated the WPS-dataset using six typical segmentation models. The models effectively learn and understand the WPS-dataset characteristics during training, resulting in high performance and accuracy in wood plate segmentation tasks. We believe that our dataset can lay a solid foundation for future research in bark removal processing and contribute to advancements in this field.
**Keywords**：bark removal processing; data acquisition; data augmentation; wood plate segmentation dataset; data validation

## 1. Introduction

Wood is a versatile natural material with extensive applications and abundant resource value, making it an indispensable component in the development of human society. Wood plays a crucial role in various aspects of human life and industrial production. Wood processing is the foundation for the widespread use of wood, involving various techniques and processes to transform raw wood into specific shapes, sizes, textures, and purposes [1]. This includes processes such as harvesting, cutting, wood plate bark removal, drying, and final product manufacturing. Wood plate bark removal, in particular, is an essential step in wood processing aimed at enhancing wood aesthetics, preventing decay and insect damage, and improving wood performance, which is critical for ensuring the quality of wood processing and its products.

Wood plate bark removal, typically, refers to the process of peeling or stripping the bark from the edges of wood plate. While logs retain their outer bark during harvesting, processing, and transportation, it is often necessary to remove this layer of bark during wood plate processing to facilitate subsequent operations and utilization. The edges and surfaces of wood plate after bark removal are smoother and more suitable for further processing and decoration. Bark removal can be carried out manually or mechanically.

Using manual methods [2] to bark removal wood is a traditional approach. This manual bark removal method is commonly used for small-scale wood processing or personal projects. And it involves using woodworking tools such as a drawknife, hammer, or hand plane to determine the optimal direction for bark removal based on the wood grain and shape. Typically, the direction parallel to the wood grain is the easiest for bark removal. Using handheld tools like a drawknife or hand plane, the technician gently scrapes off the bark from the edge of the wood plate along the determined direction. The manual bark removal method requires technicians with mature skills and ample patience to accurately remove the bark from the edge of the wood plate while minimizing waste. However, the manual bark removal method has low automation and production efficiency, making it unsuitable for large-scale wood processing demands.

Mechanical bark removal [3] is an automated method, which is more suitable for large-scale wood processing compared to manual bark removal methods. This process utilizes machinery composed of mechanical components, saw blades, motors, conveyors, and simple control buttons. Operators adjust the bark removal equipment parameters based on the size and requirements of the wood plate, then place the wood plate onto the conveyor of the bark removal equipment for edge bark removal and standardized trimming. Throughout the process, operators must assess how to segment the bark of the wood plate to obtain the largest usable wood plate, relying on their visual judgment and extensive experience. Additionally, the positioning of the wood plate on the conveyor must be relatively parallel to the saw blade's position for effective trimming, requiring skillful adjustment of wood bark removal equipment parameters. The operators of such machinery undergo extensive training or practical experience to be competent in this role. The wood bark removal equipment can process approximately 10-15 pieces of wood plate per minute. While this method significantly improves efficiency compared to the manual bark removal method, it may still fall short of the actual demands for bark removal processing. Moreover, the subjective factors of manual operation significantly affect the quality of wood plate bark removal, such as misalignment between the placement of wood plate and the saw blade, and inaccurate judgment of the wood plate size. These issues can result in significant wood resource wastage and residual bark, impacting the quality of the bark removalled wood plate and subsequent processing.

Researchers addressing the challenges of mechanical bark removal have introduced advancements in machine vision technology [4]. Typically, this method involves using cameras or other image acquisition devices on the wood bark removal equipment to capture images of wood plates. The process includes several stages such as preprocessing, image segmentation, non-wood region identification, wood region segmentation, and post-processing to identify and segment the wood plate area. In the preprocessing stage, the collected images of wood plates undergo various operations, such as noise reduction, smoothing, grayscale conversion, etc., to enhance image quality and reduce noise. In the image segmentation stage, various image segmentation algorithms such as thresholding [5], region growing [6], and watershed segmentation [7] are employed to divide the image into different regions. These regions typically include bark, wood plate, and background. In the feature extraction stage, features such as color, texture, shape, and other relevant characteristics are extracted from each segmented region obtained from the previous step of image segmentation. These features provide quantitative representations of the image content, which are used for subsequent analysis and decision-making. After feature extraction, the process proceeds to non-wood region identification. Based on the extracted features, classification algorithms or clustering methods are employed to identify non-wood regions. This is typically treated as a binary classification problem, where each region is labeled as either wood plate or non-wood plate (comprising bark and background). Next, the wood plate region segmentation stage involves separating the regions identified as non-wood regions from the rest of the image. This results in the final segmentation of the wood plate regions. Finally, the post-processing stage involves refining the segmentation results to improve accuracy. This includes tasks such as noise removal, connecting fragmented regions, and adjusting region boundaries. These post-processing steps ensure that the segmented wood plate regions are accurately delineated and ready for further analysis or downstream processing in wood processing workflows.

By following the above steps, it is possible to achieve the recognition and segmentation of wood plate regions in the images, quickly and effectively extracting the wood plate parts from the images. This approach eliminates the subjective influence of manual factors on the quality of wood plate bark removal, significantly improving the quality and efficiency of bark removal while reducing unnecessary waste. Although the direct method of recognizing and segmenting wood plate regions from the images is straightforward, it also has several drawbacks and limitations:

(1) Diversity in wood plate regions: wood plate regions in the images can exhibit significant diversity due to variations in backgrounds, lighting conditions, occlusions, and other factors, which complicates the recognition and segmentation process.

(2) Sensitivity to parameters and thresholds: the segmentation algorithms often require setting parameters or thresholds for segmentation, and the choice of these parameters can greatly impact the results, necessitating tuning and optimization.

(3) Manual feature design requirement: the methods may require manual feature design and selection to describe non-bark regions. This demands a deep understanding and experience in image processing and feature engineering, and may not comprehensively capture all region characteristics.

(4) Accuracy limitations: Directly identifying non-bark areas may be affected by interference from other objects in the image, leading to reduced accuracy and incomplete segmentation of target areas.

(5) Challenges with complex backgrounds: dealing with complex and varied backgrounds, such as those with intricate textures, multiple objects, or occlusions, can pose challenges to the accuracy and robustness of directly identifying wood plate regions.

With the wide application of deep learning-based algorithms in the field of machine vision, especially in the field of target recognition and defect detection, a practical solution is provided to address the above challenges. Many Frameworks, based on different kinds of CNN models, such as CNN [8], SSD [9], Faster-CNN [10,11], Mask R-CNN [12,13], etc., have been derived to be applied to wood surface defect recognition. YOLO family of models [14,15] and their improved versions [16,17] as well. To further enable the application of deep learning-based algorithms, a large-scale image dataset of wood surface defects is also proposed for vision-based automatic quality control processes [18]. In addition, numerous aspects of the wood industry are experimenting with research using deep learning-based algorithms, for example, species identification [19], wood NIR classification [20], wood segmentation [21], and Wood Stiffness Prediction [22], etc..

Inspired by the above research, this paper attempts to apply deep learning-based algorithms to wood plate bark removal processing as a way to address the issues and break the limitations of traditional machine vision methods. Compared with traditional machine vision methods, the deep learning-based method can be trained directly on the wood plate image with bark, without the need to manually carry out too much image preprocessing, and simplify the process of image segmentation, feature extraction and region identification. Meanwhile, the deep learning-based method only needs to learn a small amount of sample data to obtain the basic features of wood plate, which provides a great possibility for wood plate segmentation in bark removal processing.

Given such considerations, we can utilize deep learning-based segmentation models (semantic segmentation or instance segmentation) to build a deep learning framework suitable for wood plate bark removal processing. Meanwhile, the training and testing of the model are inseparable from the dataset. We equally need a generalized dataset to complete the model training and testing study. In this way, the wood bark removal equipment configured with the deep learning model will greatly improve the existing wood plate bark removal processing, reduce the labor cost, enhance the efficiency and product quality, and greatly avoid the waste of wood resources. However, through the review of related literature, we found that research based on deep learning methods in the field of wood plate bark removal processing is almost impossible to find, and there is no corresponding dataset available.

In summary, in order to enable the application of deep learning-based algorithms to wood bark removal equipment, the challenge that needs to be solved is the construction of the dataset. Therefore, an experiment was conducted in a real bark removal processing environment with the aim of obtaining a wood board segmentation dataset named WPS-dataset (wood plate segmentation dataset in bark removal processing). The industrial environment allowed us to acquire a large amount of authentic data from the production line. In this experiment, we obtained 1772 raw data samples of wood plates with bark using the designed wood plate image acquisition device. We filtered the obtained raw data samples, resulting in a total of 1621 high quality raw data samples. Data enhancement was used to expand the raw data samples to 4863. Then, with the help of image annotation software LabelMe [23], we annotated the bark segmentation of wood plate to obtain the corresponding masks, and randomly divided the data set into training set and validation set according to the ratio of 8:2 for semantic segmentation model training and validation. In addition, we used several semantic segmentation models to conduct semantic segmentation experiments, which were used to verify the validity and reliability of the WPS-dataset.

This paper is divided into the following main sections. In Section Ⅱ, we describe in detail the device for capturing plank images, the process of making the WPS-dataset including data augmentation, annotation, and data logging and segmentation. In Section III, six semantic segmentation models are used for semantic segmentation experiments using the WPS-dataset, respectively. The performance of the WPS-dataset is evaluated. In Section IV, the study is summarized and discussed.

The main contributions of our research work are summarized as follows:

(1) To the best of our knowledge, the proposed WPS-dataset is the first comprehensive dataset designed specifically for wood bark removal processing.

(2) We carefully evaluated the performance of the WPS-dataset using widely-used semantic segmentation models and standard evaluation metrics for datasets, and we obtained compelling evidence of its effectiveness.

(3) Using the WPS-dataset to train segmentation models for the wood bark removal equipment enables accurate and reliable completion of wood bark removal tasks. This approach holds promising applications in the field of wood processing.

## 2. Materials and methods

Since the experiment is being conducted in a real industrial environment, the primary issue to address first is the collection of wood plate image data. An important consideration is that the data collection process must account for the operation of the wood bark removal equipment, ensuring that the experiment is conducted while maximizing its profitability. Additionally, the conveyor belts of the wood bark removal equipment operate at high speeds, making manual image capture methods impractical. Therefore, our main goal is to develop a device capable of automating data collection in a real industrial environment. The entire process of acquisition, including the postprocessing steps, is depicted in Figure 1.

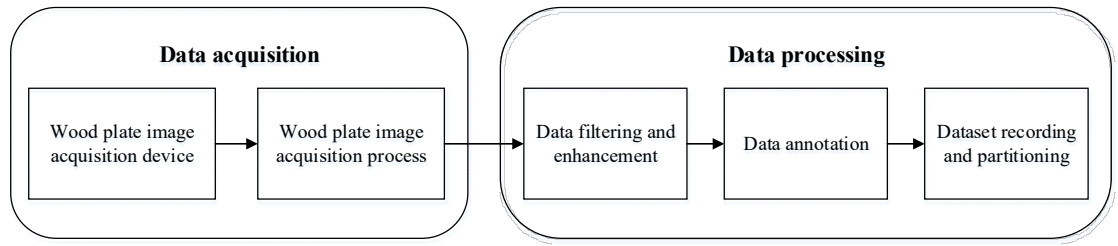

**Figure 1.** Research stages, including data acquisition and data processing steps.

### 2.1 Data acquisition
#### 2.1.1 Wood plate image acquisition device

Based on the wood bark removal equipment used in real industrial environments, this study has developed a wood plate image acquisition device that can be assembled onto such equipment, as shown in Figure 2. The device consists of a mechanical frame, a camera, several light sources, and several photoelectric sensors. The mechanical frame is used to mount the camera and light sources and is fixed above the wood bark removal equipment. To ensure image quality, we selected the Dahua A3600CU60 color area scan industrial camera, which has a sampling frequency of 60Hz and a resolution of 3072×2048, meeting the requirements for image acquisition efficiency and clarity. The camera is positioned at a height of 0.62m above the horizontal plane of the wood plate image capture area, allowing for complete capture of wood plate images ranging from 0.50m to 0.60m in length. The light sources are positioned at a height of 0.56m above the horizontal plane of the image capture area, providing specific illumination to the wood plate, resisting environmental light interference, and ensuring image stability. The photoelectric sensors used are Omron E3S-GS3E4fixed respectively on the clamp rollers within the capture area. These sensors have a detection distance of up to 30mm, ensuring that the boards with bark remain accurately positioned in the wood plate image capture area.

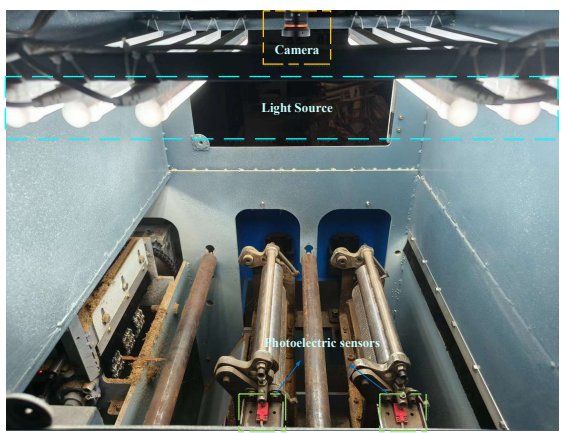
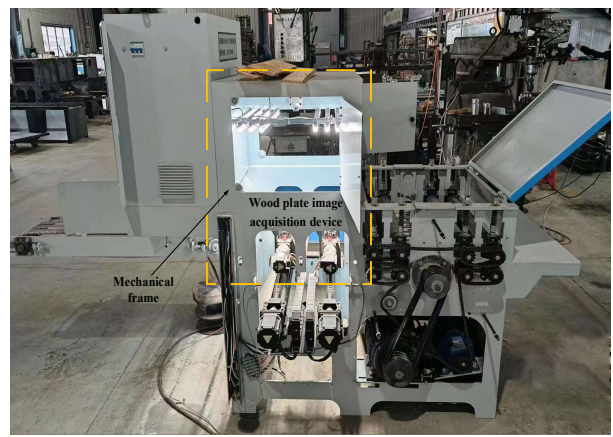

(a) inside wood plate image acquisition device    (b) assembled wood bark removal equipment

**Figure. 2** Wood plate image acquisition device.

### 2.1.2 Wood plate image acquisition process

After installing the designed wood plate image acquisition device onto the wood bark removal equipment, to obtain the image data of the boards, the process is as follows:

Step1: A wood plate is manually placed onto the conveyor belt. When the conveyor belt starts operating, the board is transported forward towards the image capture area. At this point, the photoelectric sensor generates a pulse on the acquisition board of the computer.

Step2: When the wood plate reaches the designated area, the camera begins scanning to capture the image of wood plate. The captured image is then transmitted to the computer via a single twisted pair Ethernet connection.

Step3: The scanning process stops when the wood plate exits the capture area.

Step4: The captured wood plate images are then processed by the camera to produce images with a resolution of 3072x2048 pixels.

Step5: Each high-resolution color image occupies approximately 15 MB of disk space. Therefore, for this experiment, two external hard drives with a capacity of 512 GB each are used to collect the image data.

To conserve CPU time, detailed online processing is not performed. Instead, the image data is saved to the hard drives for subsequent processing. This approach allows for efficient and systematic collection of high-quality image data from the wood plates during the bark removal processing, facilitating further analysis and experimentation.

### 2.2 Data processing
### 2.2.1 Data filtering and enhancement

During a data collection process lasting several hours, we obtained a total of 1772 images of wood plate. To improve the efficiency of image acquisition in a high-speed environment, where the process is continuous, challenges such as empty conveyances or overlapping board conveyances can lead to the capture of images that do not meet annotation requirements. After manually removing some blurry, low-quality, and obscured images, the dataset was reduced to 1621 images. To expand the dataset, we employed data augmentation techniques by rotating, horizontally flipping, and vertically mirroring the original images, thereby expanding the dataset to 4863 images. Through data augmentation, we increased the diversity of training data for wood plate, making it more representative and enhancing the model's generalization ability while reducing the risk of overfitting. Moreover, to address potential issues arising from insufficiently capturing the variability introduced by data augmentation, we conducted multiple rounds of wood plate image acquisition by varying work scenarios, adjusting camera brightness, and changing object conveyance angles. This approach better covers the true distribution of the data, reducing biases introduced by data augmentation and enhancing image variability. Figure 3 depicts examples of data augmentation operations applied to two different sets of wood plate images.

| original image | rotating | horizontally flipping | vertically mirroring |

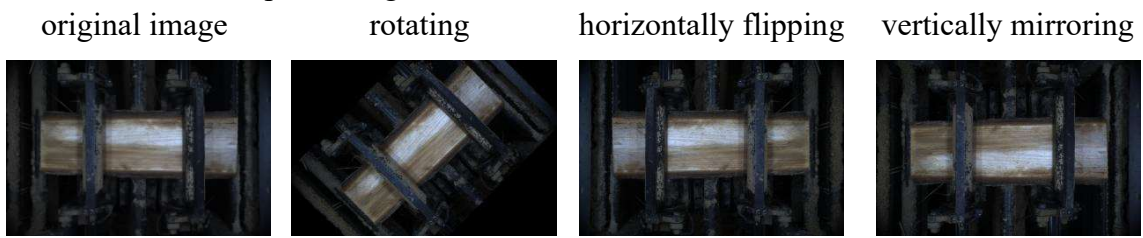

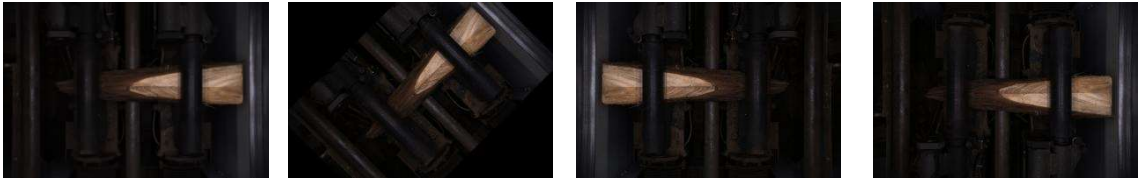

**Figure 3.** Comparison of data augmentation effects between two groups. From left to right are the original image, rotated image, horizontally flipped image, and vertically mirrored image.

**2.2.2 Data annotation**

The annotation work in this study was primarily conducted using LabelMe. LabelMe is an annotation tool for image annotation that allows annotations to be made anywhere and shared instantly. Compared to other annotation tools, LabelMe has a more intuitive interface and operation, making it easy for users to perform annotation tasks. Users can visualize annotated data during the annotation process, showing real-time annotation results, which allows for quick inspection and adjustment of annotations to ensure accuracy and consistency. Additionally, LabelMe supports exporting annotation data in JSON format, which contains detailed descriptions of the images and annotations. This enables users to conveniently integrate annotated data with other machine learning or deep learning frameworks [24].

In this work, to ensure the rigor of the annotation task, the dataset annotation was completed by two researchers. One researcher annotated the dataset using LabelMe, independently performing the image annotation process, while the other researcher was responsible for inspecting the annotation standards and correcting any non-standard image annotations.

Furthermore, LabelMe supports various annotation shapes such as polygons, rectangles, circles, polylines, line segments, and points (suitable for tasks like object detection and image segmentation). After considering the impact of annotation methods on annotation efficiency for our dataset, we decided to use the polygon annotation method to complete the annotation process, as shown in Figure 4. We manually drew desired board regions on the wood plate images using the polygon drawing button on the toolbar and added labels to the enclosed regions using the label panel on the right. The annotation information for each processed image in the dataset was then saved as a JSON file.

Compared to other annotation methods, creating annotations with polygons can eliminate background pixels, accurately describe object shapes and sizes, capture precise dimensions of objects, and annotate object curves and different angles, thereby improving the accuracy of model training.

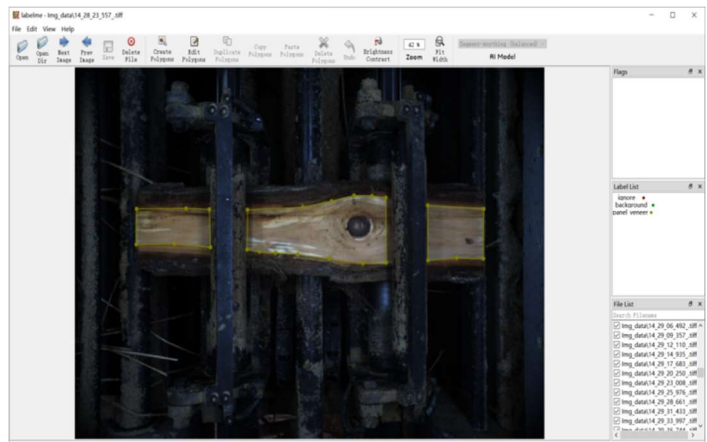

**Figure 4.** Using LabelMe for image annotation.

**2.2.3 dataset recording and partitioning**

After completing the annotation process, we constructed a semantic segmentation dataset containing 3573 images of wood plate, which we named "Img_data." Additionally, we defined a custom semantic segmentation label named "panel_veneer," which is saved in a text file named "label.txt." Figure 5 shows examples of images from the semantic segmentation dataset. Next, we randomly divided the dataset into training and validation sets using an 8:2 ratio. Both sets include original images and corresponding annotation information. Specifically, the training set consists of 3890 images, and the validation set consists of 973 images. These sets are used for model training and validation purposes.

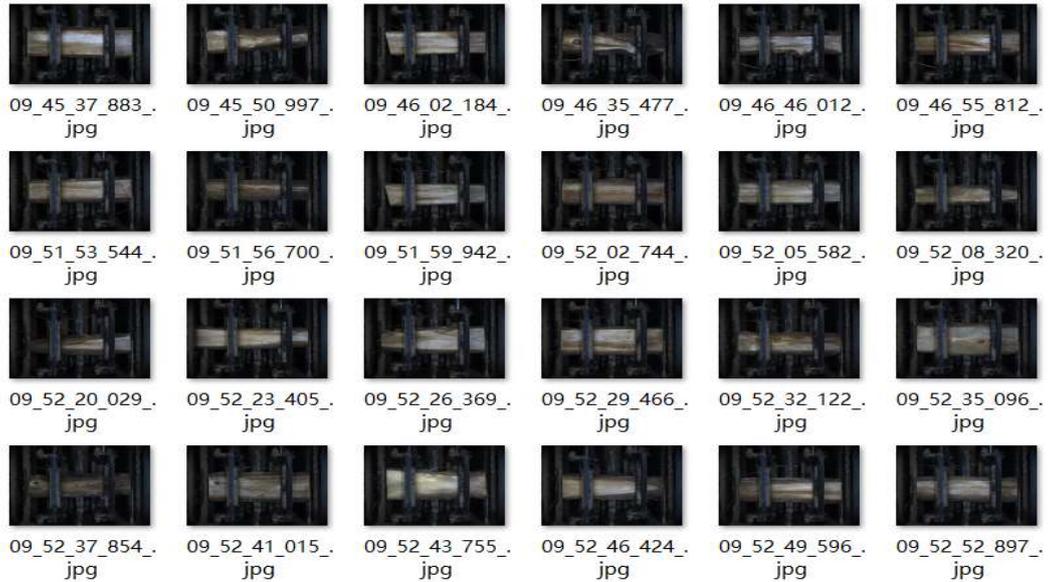

(a) original images

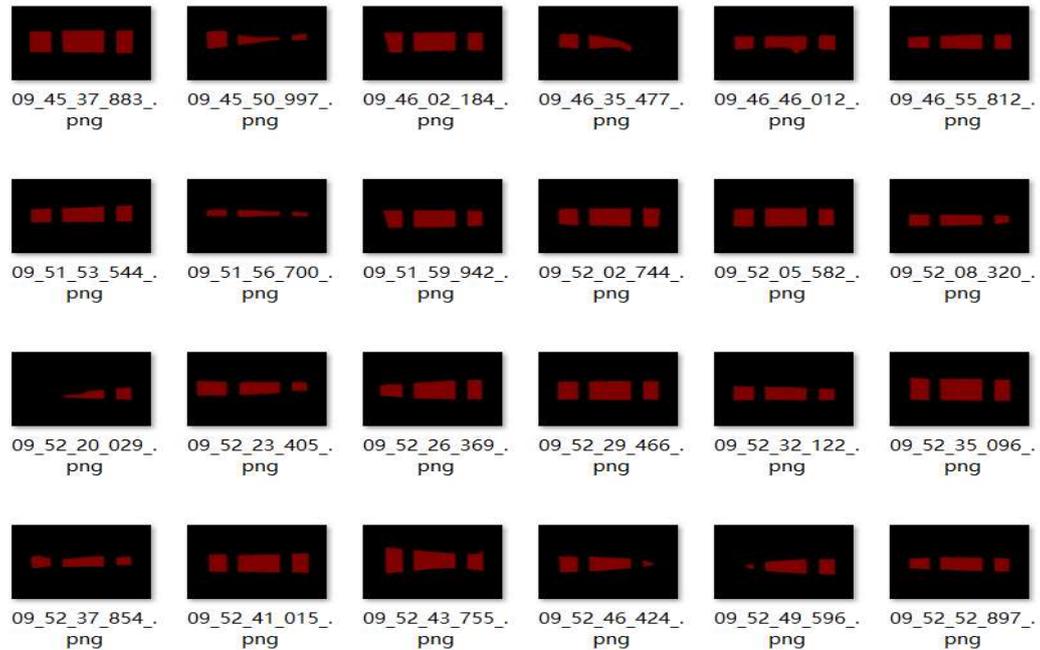

(b) Ground Truth

**Figure 5.** Partial WPS-dataset.

## 3. Technical validation

The technical validation of the dataset was conducted by assessing the quality of the assigned labels by employing widely used semantic segmentation models. For this purpose, we utilized six typical deep learning-based semantic segmentation models: Fully Convolutional Network (FCN), U-Net, Pyramid Scene Parsing Network (PSPNet), High-Resolution Network (HRNet), DeeplabV3, and DeeplabV3+.

Among them, FCN [25] uses convolutional neural networks to transform from image pixels to pixel categories. It can accept input images of arbitrary sizes without requiring all training and testing images to have the same dimensions, thus avoiding issues related to repetitive storage and computation of convolutions using pixel blocks. Built on an encoder-decoder architecture, U-Net [26] supports training with a small amount of data and can segment each pixel to achieve higher segmentation accuracy. PSPNet [27] uses multi-scale feature fusion to enhance global contextual information. By applying pyramid pooling modules to the semantic segmentation field, it can better learn global contextual information of scenes and effectively improve segmentation accuracy. HRNet [28] maintains high-resolution representations throughout the process by parallel connections of high-resolution and low-resolution convolutional layers. Instead of restoring resolution from low to high, it performs repeated multi-scale fusion with help from low-resolution representations of similar depth and level to enhance high-resolution representations. DeeplabV3 [29] optimizes the structure of Atrous Spatial Pyramid Pooling (ASPP) using dilated convolutions with different rates and Batch Normalization) (BN layers. It arranges ASPP modules in a cascaded or parallel manner and integrates image-level features into ASPP modules to better capture object and region details as well as contextual information. DeeplabV3+ [30] employs Atrous Spatial Pyramid Pooling (ASPP) to better extract features under different conditions. Additionally, inspired by the structure of U-Net, it adds an upsampling decoder module to optimize edge precision, achieving good performance.

### 3.1 Evaluation metrics

To assess the accuracy of semantic segmentation models when using WPS-dataset, and simultaneously validate the technical utility of the WPS-dataset. Our experiment employed five evaluation metrics: Mean Intersection over Union (MIoU) [31], Accuracy, Precision, Recall, and F1-score. The practicality of the WPS-dataset was validated through comparative analysis of the performance of the six aforementioned network models. The formulas for each evaluation metric are as follows:

MIoU measures the average intersection over union for all classes. It is calculated as:

$$\text{MIoU} = \frac{1}{k+1} \sum_{i=0}^{k} \frac{\text{TP}_i}{\text{TP}_i + \text{FP}_i + \text{FN}_i} \quad (1)$$

Where, $k$ represents the number of classes, $k+1$ represents the number of classes plus the background, and $i$ represents one specific class. Assuming there are $k+1$ (0, … $k$) classes in the dataset, where $k=0$ typically represents the background. In our experiment, the value of $k$ for the WPS-dataset is 1. $\text{TP}_i$ is the true positive for class $i$, $\text{FP}_i$ is the false positive for class $i$, $\text{TN}_i$ is the false negative for class $i$.

Accuracy measures the overall accuracy of the segmentation. It is calculated as:

$$\text{Accuracy} = \frac{\text{TP} + \text{TN}}{\text{TP} + \text{TN} + \text{FP} + \text{FN}} \quad (2)$$

Where, where TP is the number of true positives, TN is the number of true negatives, FP is the number of false positives, and FN is the number of false negatives.

Precision measures the proportion of true positive predictions out of all positive predictions. It is calculated as:

$$\text{Precision} = \frac{TP}{TP + FP} \qquad (3)$$

Recall measures the proportion of true positive predictions out of all actual positives. It is calculated as:

$$\text{Recall} = \frac{TP}{TP + FN} \qquad (4)$$

F1-score is the harmonic mean of precision and recall, providing a balanced measure between the two. It is calculated as:

$$F1 = 2 \times \frac{\text{Precision} \times \text{Recall}}{\text{Precision} + \text{Recall}} \qquad (5)$$

These metrics allow for a comprehensive evaluation of semantic segmentation models based on their predictive performance across multiple classes. Comparing these metrics among different network models helps assess the dataset's utility and the effectiveness of the models in handling semantic segmentation tasks.

### 3.2 Semantic segmentation model training

To meet the training requirements of the semantic segmentation models, in the experimental setup, we equipped a computer with Windows 10 operating system, an Intel(R) Core(TM) i5-9300 CPU @ 2.40 GHz processor, and an NVIDIA GeForce GTX 1650 graphics card. The training environment for the six aforementioned network models (FCN, U-Net, PSPNet, HRNet, DeeplabV3, and DeeplabV3+) used PyTorch version 1.9.1.

The specific training process is as follows:

(1) Data Preparation: Before training, original images and label images were placed in the folders JPEG Images and Segmentation Class under the VOC2007 directory in the VOCdevkit folder.

(2) Parameter Settings: Detection categories (num classes), learning rate, batch size, and epochs were configured. Since this experiment aimed to segment the wood parts in wood plate images, the dataset images were divided into wood and non-wood regions, hence setting num classes to 2. To ensure both training accuracy and efficiency given the computer's performance, we set the learning rate to 1e-4, batch size to 4, and epochs to 100.

(3) Downloading Pretrained Weight Files: Pretrained weight files for the FCN and DeeplabV3 models required resnet50 weights, the U-Net model required vgg weights, PSPNet and DeeplabV3+ models required mobilenetv2 weights, and the HRNet model required hrnetv2_w18 weights, which were downloaded from the official websites.

(4) Optimizer Selection: To reduce training time and improve model convergence speed, we used the SGD optimizer, which calculates gradients using a single sample per update, allowing the model to converge quickly during training.

### 3.3 Analysis of experimental results

In our experiment, we performed wood plate segmentation on the WPS-dataset using the six aforementioned network models. We evaluated the experimental results using mentioned performance metrics in 3.1. The evaluation results are shown in Table 1.

**Table 1.** Evaluation metrics for experimental results of the WPS-dataset

| Network model | MIoU | Accuracy | Precision | Recall | F1 |
|---|---|---|---|---|---|
| FCN | 0.9694 | 0.9931 | 0.9839 | 0.9872 | 0.9855 |
| U-Net | 0.9824 | 0.9960 | 0.9903 | 0.9919 | 0.9911 |
| PSPNet | 0.9629 | 0.9916 | 0.9829 | 0.9790 | 0.9809 |
| HRNet | 0.9779 | 0.9950 | 0.9872 | 0.9903 | 0.9887 |
| DeeplabV3 | 0.9686 | 0.9930 | 0.9836 | 0.9869 | 0.9852 |
| DeeplabV3+ | 0.9738 | 0.9942 | 0.9850 | 0.9882 | 0.9866 |

Based on the data presented in Table 1 for the evaluated models, we can analyze the performance and characteristics of the dataset as follows:

(1) The WPS-dataset achieves a high MIoU across all listed models, ranging from 0.9629 to 0.9824. This indicates accurate segmentation of categories within the WPS-dataset, with models effectively capturing intersections and unions between different categories of pixels.

(2) All models exhibit high accuracy, ranging from 0.9916 to 0.9960. This suggests precise overall pixel classification within the WPS-dataset, where models accurately predict the majority of pixel categories.

(3) There are slight differences in precision and recall among different models, but overall performance is stable and efficient. all listed models demonstrate good precision and recall, indicating accurate prediction of positives and coverage of most true positives.

(4) The F1-score combines precision and recall to provide a comprehensive evaluation of model prediction effectiveness. all listed models also perform well in F1-score, demonstrating effective balancing of precision and recall during predictions.

In summary, based on these evaluation metrics, the WPS-dataset performs well in wood plate segmentation tasks, showing clear category distinctions and accurate pixel classification results. Models effectively learn and understand WPS-dataset characteristics during training, leading to high performance and accuracy in wood plate segmentation tasks.

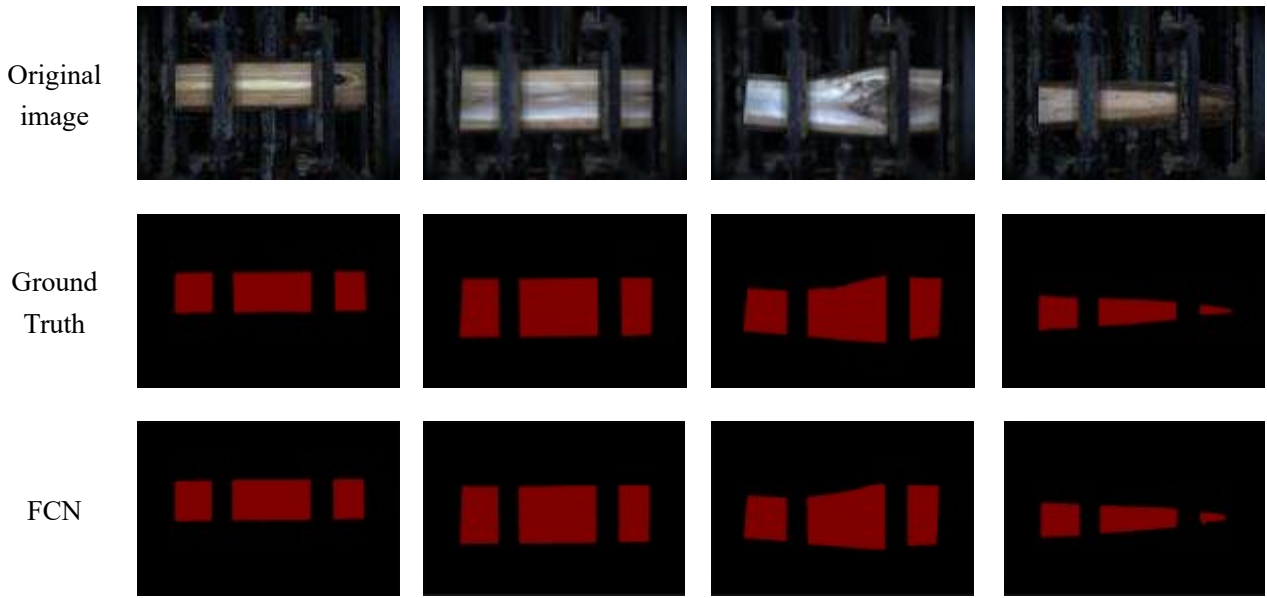

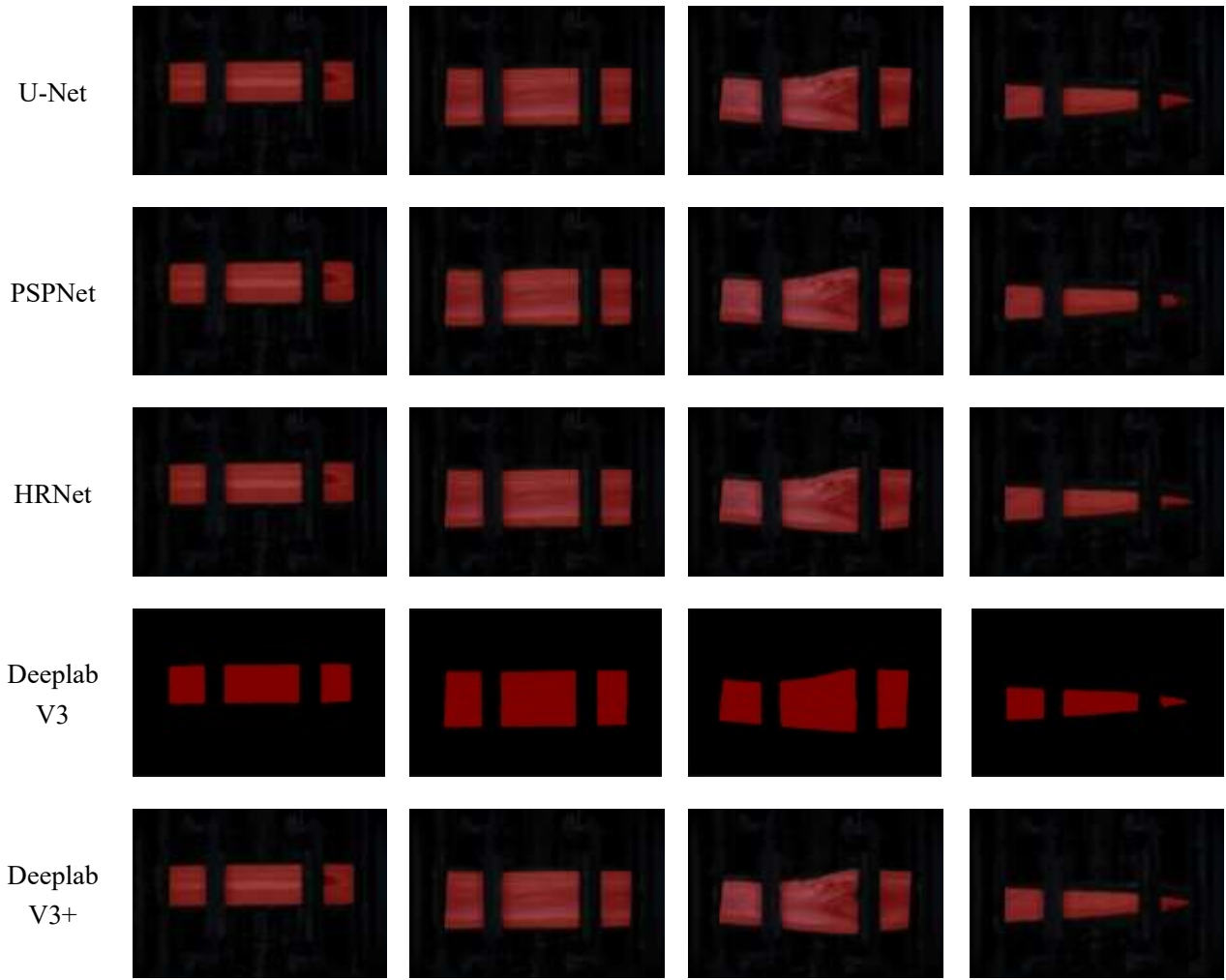

**Figure 6.** Partial visualization results of the predicted masks.

To visually demonstrate and validate the usability of the WPS-dataset, we compared the predicted masks from all listed models with the Ground Truth through visualization. Due to the large number of predicted images of wooden plates, we selected only four sets of wood plate images with different shapes as examples. Figure 6 shows the predicted masks obtained from experiments on the WPS-dataset. From Figure 6, it can be observed that all listed models used in the experiment accurately segmented the wood plate regions from the non-wood plate regions (bark and background). The predicted masks cover the target objects, meaning they cover most of the pixels in the wood plate regions while maintaining pixel classification consistency and continuity. Additionally, the predicted masks are quite similar to the Ground Truth with minimal differences.

## 4. Conclusions and discussion

The bark removal process of wood plate is an important stage in the wood industry. However, existing bark removal equipment often suffers from poor efficiency, low effectiveness, and significant waste of wood resources. This article explores the potential practical application of deep learning-based semantic segmentation models in the context of bark removal processing. To address this, we propose a benchmark dataset for wood plate segmentation in bark removal processing named WPS-dataset, aiming to provide a universal dataset that enables the application of deep learning-based semantic segmentation models.

To obtain real wood plate image data in an industrial environment, we designed a specialized device for capturing wood plate images and installed it on the bark removal equipment. This allowed us to collect wood plate image data in real industrial settings. After data filtering, augmentation, annotation, and partitioning steps, we compiled a benchmark dataset containing 3573 images. We conducted technical verification experiments using six typical semantic segmentation models. The experimental results demonstrate the general applicability of WPS-dataset, as the models effectively learned and understood the characteristics of the WPS-dataset during training. The WPS-dataset performs well in wood plate segmentation tasks.

Our research marks the first step in applying deep learning-based algorithms to bark removal processing in the wood industry. This work serves as a valuable reference and holds broad application prospects for researchers, engineers, and businesses in this field. Despite the promising practicality demonstrated by our dataset in the technical verification, there are still some issues that require further research and validation. For example, the WPS-dataset may suffer from sample imbalance, with most images being of regular-shaped wood plates and fewer images of uniquely shaped plates. The WPS-dataset may lack diversity and representativeness, failing to cover various changes and scenarios found in real industrial settings. To address this, we need to enhance dataset diversity by including different angles, lighting conditions, backgrounds, and environmental factors to improve model generalization. Additionally, the WPS-dataset establishment process should be continuous and maintained, with ongoing additions of new data samples, updates to labels, and improvements in data quality to ensure the dataset remains effective and practical. This will guide our future research efforts in this direction.

**Author Contributions:** Rijun Wang: Supervision, Writing-review & editing, Funding acquisition, Investigation, Methodology. Guanghao Zhang: Writing-original draft, Software, Methodology. Fu-long Liang: Software, Methodology, Data curation. Bo WANG: Conceptualization, Validation, Data curation. Xiangwei Mou: Resources, Writing-review, Funding acquisition. Yesheng Chen: Data curation, Software. Peng Sun: Funding acquisition, Conceptualization, Validation. Canjin Wang: Methodology, Supervision, Writing-review & editing.

**Funding:** This study was co-supported by the Science and Technology Planning Project of Guangxi Province, China (No. 2022AC21012); the industry-university-research innovation fund projects of China University in 2021 (No. 2021ITA10018); the fund project of the Key Laboratory of AI and Information Processing (No. 2022GXZDSY101); the Natural Science Foundation Project of Guangxi, China (No. 2018GXNSFAA050026); National Natural Science Foundation of China(No. 62105305), Fundamental Research Program of Shanxi Province(No. 20210302123068).

**Data Availability Statement:** The data that support the findings of this study are available from the corresponding author, Rijun WANG, upon reasonable request.

**Conflicts of Interest:** The authors declare no conflict of interest.